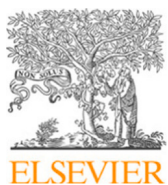
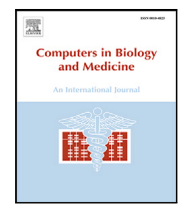

# Automated detection of pediatric congenital heart disease from phonocardiograms using deep and handcrafted feature fusion

Abdul Jabbar [a],[*], Ethan Grooby [a],[b], Yang Yi Poh [a], Khawza I. Ahmad [c], Md Hassanuzzaman [e], Raqibul Mostafa [c], Ahsan H. Khandoker [d], Faezeh Marzbanrad [a]

[a] *Electrical and Computer System Engineering, Monash University, Clayton, Melbourne, 3800, VIC, Australia*
[b] *Biomedical Engineering, McGill University, Quebec, Canada*
[c] *Electrical and Electronic Engineering (EEE), United International University, Dhaka, Bangladesh*
[d] *Healthcare Engineering Innovation Group (HEIG), Department of Biomedical Engineering and Biotechnology, Khalifa University, Abu Dhabi 127788, United Arab Emirates*
[e] *Department of Electrical and Computer Engineering (ECE), Duke University, USA*

A B S T R A C T

Congenital heart disease (CHD) is the most common type of birth defect, impacting about 1% of live births worldwide. Echocardiography, the gold-standard diagnostic method, is costly and inaccessible in low-resource settings. Diagnosis is delayed due to limited skilled experts, whose ability to interpret pathological patterns varies significantly, causing inter- and intra-clinician variability. Therefore, we present a new method for a more accessible diagnostic modality, the digital stethoscope, to detect CHDs. Our method is based on deep feature fusion, integrating deep and handcrafted features for the automated early detection of CHDs. For this work, Phonocardiography (PCG) recordings were obtained from 751 pediatric subjects (Age:1 month- 16 years) in Bangladesh, ranging from infants to adults at four auscultation locations: mitral valve (MV), aortic valve (AV), pulmonary valve (PV), and tricuspid valve (TV). These recordings were labeled based on confirmed diagnoses by cardiologists as either cases of CHD or non-CHD. The results demonstrated that our proposed model achieved an accuracy of 92%, a sensitivity of 91%, and a specificity of 91%, based on a patient-wise split of 70% training, 20% validation, and 10% testing. Furthermore, the Area Under the Receiver Operating Characteristic curve (AUROC) of 96%, and an F1-score of 92%. This model promises efficient real-time remote detection of CHDs as a cost-effective screening tool for low-resource settings.

## 1. Introduction

Heart sounds, particularly in children, are valuable and accessible sources of information to monitor and diagnose cardiac problems. These sounds, with S1 and S2, as shown in Fig. 1, being the most noticeable, are produced by the heart chambers that pump blood and the heart valves closing. Using a PCG signal generated by digital stethoscopes or sensitive microphones, these sounds can be recorded and digitally processed, aiding in detailed analysis. This non-invasive method assists medical professionals in assessing the heart's condition by examining the timing, intensity, and characteristics of these sounds [1]. While heart auscultation is an accessible method for CHD screening, it poses challenges in clinical practice, especially with younger patients. It demands considerable skill and experience, and several factors can complicate the auscultation. Heart sounds and murmurs are often faint and consist of low-frequency elements crucial for identifying irregularities. Furthermore, methods designed for adults may not always be effective for children, whose heart rates differ. Children's heart sounds are frequently contaminated by background noise, including breathing, crying, environmental sounds, and movement artifacts. Moreover, cardiac disorders in children can manifest in diverse ways, requiring specific knowledge and tools for accurate diagnosis [2].

Screening children for CHDs is one of the most important aspects of neonatal and pediatric care. CHDs refer to structural anomalies in the heart present at birth, ranging from minor to severe, affecting its anatomy, function, or both. Its prevalence varies across different demographics and regions. CHD rates are more prevalent in low- and middle-income countries than in high-income ones. For example, in the United States, an incidence rate of 6.9 per 1,000 live births has been reported, whereas in India, the incidence rate was found to be 8.2 per 1,000 live births [3,4]. This highlights a greater need for accessible and cost-effective CHD screening methods to ensure that more newborns, regardless of their place of birth and their socioeconomic background, can receive early diagnosis and life-saving treatment. This approach will reduce the long-term medical expenses associated with untreated CHD, benefiting both patients and the larger healthcare system.

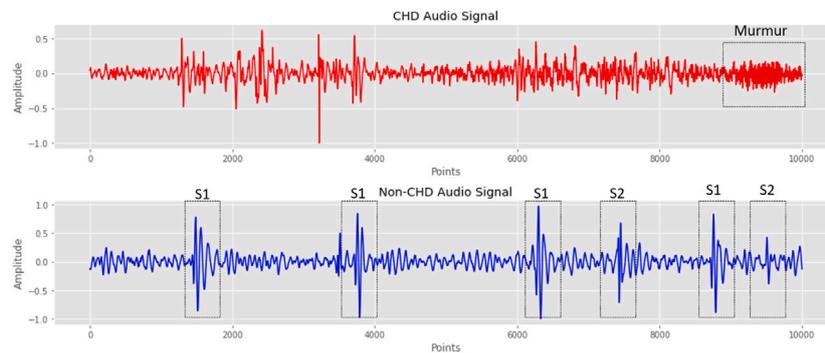

**Fig. 1.** Audio signal for Congenital heart disease (CHD) patient and healthy child.

Certain critical heart conditions, such as hypoplastic, left heart syndrome, transposition of the great arteries, and critical aortic stenosis, can be lethal in the initial days or weeks of life, with the highest mortality rates occurring during these early periods [5,6]. Additionally, the long-term implications of CHD contribute to increased mortality rates. Individuals with CHD are at a greater risk of developing complications, including arrhythmias, endocarditis, and heart failure, which can notably reduce their life expectancy [7].

Early diagnosis of CHD is crucial for several reasons. Firstly, it enables the timely implementation of appropriate management strategies and interventions. Advances in surgical techniques and medical technology have significantly enhanced the outlook for newborns and children suffering from congenital heart defects. Early detection and treatment of severe CHDs in children have been shown to significantly improve survival rates and reduce the need for emergency treatments [8]. Secondly, early diagnosis facilitates the prevention of CHD-related complications. Prompt intervention can prevent or minimize the evolution of issues such as heart failure and pulmonary hypertension, which may arise from untreated or inadequately managed congenital heart defects. Finally, early diagnosis improves the overall quality of life of people with CHD. Early interventions can foster physical and mental growth, reduce symptoms, and improve cardiac function. Furthermore, it provides families with access to suitable therapy and support, helping them manage the challenges associated with CHD [8].

To address these challenges, there is a pressing need for accurate, accessible, and affordable CHD screening tools that can overcome limitations posed by noisy environments, pediatric-specific characteristics, and the limited availability of specialists, particularly in LMICs. In this context, we propose a hybrid deep learning model for CHD screening using digital stethoscopes as affordable and practical tools, particularly in low-resource environments. Our model is innovative in its integration of both handcrafted and deep features, enabling it to leverage the strengths of both manual feature extraction and the complex pattern recognition capabilities of deep learning. The hand-selected features of our model, such as features in the frequency domain, features in the time domain, and statistical measures, are clinically relevant because they reflect the underlying abnormalities in heart sounds associated with CHD. These features capture subtle variations in heart sounds that may indicate the presence of murmurs or irregular heartbeats, which are often key indicators of CHD. The choice of these features is based on existing literature [9,10], which demonstrates their importance in distinguishing between normal and abnormal heart conditions. Moreover, these features can be easily extracted from PCG signals, making the model applicable in real-world settings, especially in LMICs where diagnostic resources are limited.

The main objectives of this study are:

- To develop a hybrid deep learning-based model capable of classifying pediatric patients into CHD and Non-CHD from PCG sounds
- To integrate both handcrafted and deep features to enhance features representation for better classification
- To evaluate the feasibility and performance of this approach in resource-constrained settings, thereby supporting early, accessible, and reliable CHD detection

The rest of the paper is structured into several sections. Section 2 presents a summary of previous work. Section 3 demonstrates the dataset and the complete methodology used in our research. Following this, Section 4 presents the performance of our proposed model, summarizes the fundamental insights, and a comparison with previous literature. Lastly, Section 5 presents the final assessment of the research.

## 2. Related works

Over the past twenty years, machine learning (ML) and deep learning (DL) have greatly improved medical diagnostics, enhancing disease classification and identification, which results in better-informed treatment options [11–14]. Various algorithms have been introduced to automate the classification of heart diseases using PCG signals, using signal processing and traditional ML algorithms. El Badlaoui et al. utilized a Support Vector Machine (SVM) in combination with Principal Component Analysis (PCA), exploring various kernel sizes and hyperparameters across two different PCG datasets. In this approach, features are extracted from both the time and frequency domains, indicating the potential for reliable detection of heart murmurs using PCG analysis [15]. Wu et al. explored the application of the Hidden Markov Model (HMM) with Mel-frequency cepstral coefficients (MFCC), investigating different hyperparameters. In this method, MFCC-based HMM's classification performance was studied with a heart sound signal by changing the model's number of mixtures, number of states, and analysis frame size during the MFCC feature extraction process [16]. Maglogiannis et al. demonstrate an automated diagnoses method that utilized integration of SVM with morphological analysis and wavelet transform to identify particular heart valve disease using heart sounds [17]. In order to diagnose CHD in pediatric patients, Burns [18] emphasizes the accuracy and sensitivity in detecting pathological murmurs and specific lesions, such as aortic stenosis and regurgitation, and Chen et al. reported on discrete time-frequency energy-based features with S-transform for heart sound classification on a private dataset [19]. Moreover, other feature extraction techniques were applied in conjunction with other PCG classifiers, including spline chirplet-based approaches, polynomial chirplet transform, and synchrosqueezing, as mentioned in [20–22].

A large public adult PCG dataset was introduced in the PhysioNet Challenge 2016 [23]. Homsi et al. the third-highest performer in the PhysioNet Challenge 2016, segmented the heart sounds into S1 and S2 and then extracted 131 features from both the time and frequency domains for the classification of heart sounds, achieving an overall accuracy of 84.48% [24]. The second-highest performer, Zabihi et al. utilized a feedforward neural network for heart sound classification without segmentation. In this approach, 40 features were extracted from the heart sounds, and 13 features were selected for classification, achieving an accuracy of 85.9% [25]. The highest performance was achieved by Potes et al. who used Adaboost and CNN for heart sound classification, extracting 124 features from the heart sounds and achieving the highest accuracy of 86.02% in the PhysioNet Challenge 2016 [26].

Additionally, numerous studies have utilized a wide array of DL methods for PCG signal classification. Li and colleagues developed a fusion system that combines deep learning characteristics extracted from PCG signals with multi-domain features [27]. Using the PhysioNet challenge 2016 dataset, Singh et al. proposed a 2D scalogram approach involving a convolutional neural network (CNN) and continuous wavelet transform (CWT) [28]. Furthermore, a convolutional recurrent neural network (RNN) was developed by Alkhodari and Fraiwan to categorize various heart valve disorders [29]. Baghel et al. introduce a six-layer CNN for categorizing various heart valve disorders. Bozkurt et al. achieved remarkable results by classifying heart sounds with an accuracy of 86.02% using a deep learning model and time-frequency characteristics on the PhysioNet 2016 challenge dataset [30]. Deng et al. propose a novel method that integrates improved MFCC features with a convolutional recurrent neural network (CRNN) for heart sound classification. Unlike conventional approaches, MFCC features are computed without segmenting the heart sound signals, allowing the method to capture dynamic characteristics effectively. These features are then processed by a CRNN, which combines the strengths of CNNs for local feature extraction and RNNs for modeling long-term dependencies. Applied to the 2016 PhysioNet/CinC Challenge database, this method achieves a remarkable accuracy of 98% for binary classification of heart sounds (normal vs. abnormal) [31].

A Pediatric PCG dataset was introduced in 2022 for the PhysioNet 2022 Challenge [32]. The top performer in the PhysioNet Challenge 2022, Yujia Xu et al. utilized a hierarchical multi-scale convolutional neural network (HMS-Net) for murmur detection and outcome classification. In this approach, convolutional features were extracted from the spectrogram on multiple scales and a hierarchical architecture was used for the fusion of these features, achieving the highest accuracy of 80.6% in murmur detection class [33]. The second-highest performer, Hui Lu et al. developed a lightweight CNN and Random Forest classifier for heart disease classification. In this approach, 128 mel spectrogram features were extracted from heart sounds and demographic features were also utilized for heart sound classification, achieving an overall accuracy of 79.1% in murmur detection class [34]. The third-highest performer, McDonald et al. used a recurrent neural network and hidden semi-Markov model (HSMM) for murmur detection in heart sounds, achieving an overall accuracy of 77.6% [35].

Additionally, some researchers have explored pediatric CHD classification using privately collected PCG datasets. Aziz et al. proposed a method combining empirical mode decomposition (EMD) with MFCCs and one-dimensional local ternary patterns (1D-LTPs) for atrial septal defect (ASD) and ventricular septal defect (VSD) detection in pediatric patients [36]. Hassanuzzaman et al. introduced a transformer-based deep learning model to classify raw, unprocessed PCG signals for CHD detection. Their method combined a one-dimensional CNN with an attention transformer, eliminating the need for segmentation or handcrafted feature extraction. The model was trained on a dataset of 484 pediatric patients (2,068 recordings) collected using an FDA-cleared digital stethoscope [37]. Furthermore, Alkahtani et al. introduced an effective method for detecting CHD in children using PCG signals and a deep neural network. Their approach employs a 1D-CNN to classify raw PCG signals into CHD and Non-CHD cases, using 583 samples collected from pediatric populations in Pakistan [38].

Despite several studies, a research gap still exists when it comes to the overall performance of CHD screening. The majority of these studies still require human interaction, such as heart sound segmentation, which is subjective and greatly affected by the initial signal quality in most cases [19,24,26,30]. In some studies, the dataset utilized for training is not large enough, making it difficult to train and evaluate its performance [15,17,37,38], and the performance of some studies is still not promising for clinical applications [28–30]. Furthermore, a number of studies used simple and carefully selected datasets that might not accurately reflect the realistic state of PCG recordings, which can often be altered by noise sources [16,27]. Additionally, certain studies primarily focus on adult patients, limiting their applicability to pediatric populations due to physiological and clinical differences [24–26,31]. Moreover, the absence of information regarding ideal parameters, their meaning, and unpredictability may create uncertainty issues for deep learning models [39]. Therefore, more advanced techniques are still required to handle complex recordings more effectively, independent of their initial quality, with reduced signal interference and improved interpretability, as well as to incorporate multi-domain features for a more thorough screening.

## 3. Methods

### 3.1. Dataset

The dataset utilized in this study was collected at Bangladesh Shishu (Children's) Hospital and National Heart Foundation Hospital & Research Institute Bangladesh. Data collection involved using the EKO DUO ECG+ Digital Stethoscope, which was notable for being the first stethoscope to receive FDA clearance. Data were obtained from both healthy individuals (Non-CHD) and those with medical conditions (CHD) in a clinical setting from 01/10/2021 to 31/01/2023. Human Ethics was approved by relevant regional medical committees, including those at Shishu (Children's) Hospital (Ethics Approval Number: Admin/1714/BSHI/2021) and the National Heart Foundation Hospital (Ethics Approval Number: NHFH & amp; RI 4-14/7/AD/1132) [1]. All samples from the National Heart Foundation Hospital & Research Institute were taken from CHD patients, as these patients visited the hospital specifically for treatment. The samples from Bangladesh Shishu (Children) Hospital & Institute includes both CHD and Non-CHD patients, as this hospital primarily handles CHD diagnosis, resulting in a bias towards CHD (61%). Furthermore, the CHD cases were classified into cyanotic and acyanotic types; however, more specific labels for CHD subtypes were not available in the dataset.

This dataset includes 3004 audio recordings from 751 subjects, captured at a frequency of 4000 Hz, with 15 s per sample duration. PCG sounds were sequentially recorded from four distinct auscultation locations: the aortic valve (AV), pulmonary valve (PV), tricuspid valve (TV), and mitral valve (MV) shown in Fig. 2. The patient population is biased towards males, with ages ranging from 5 to 17 months and a CHD:non-CHD ratio of 63:37. Clinical experts classified all patients into two groups: CHD and non-CHD using CT scan, chest x-ray, 12 lead ECG, echocardiography, and clinical verification such as auscultation of the heart sounds. Tables 1–3 provide a detailed description of the dataset. Fig. 1 illustrates the sample sounds for a CHD patient and a healthy child.

### 3.2. Preprocessing

#### 3.2.1. Background noise removal

In this study, a 4th order Butterworth band-pass filter ranging from 25 to 400 Hz is applied to the audio files to eliminate high-frequency noisy components, as most heart sounds lie within this frequency range [31].

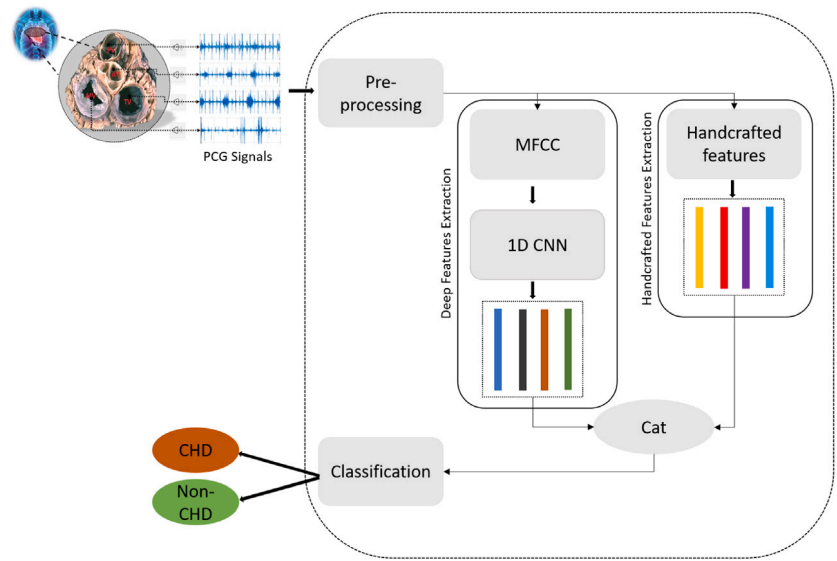

**Fig. 2.** Block diagram of the proposed multi-feature fusion model. The first phase involves pre-processing, followed by two parallel paths for feature extraction: one for deep features and another for handcrafted features. The extracted features are then fused using concatenation and passed to a classifier for final prediction.

**Table 1**
Gender distribution.

|        | CHD, (%)  | Non-CHD, (%) | Total, (%) |
|--------|-----------|--------------|------------|
| Male   | 257 (56)  | 190 (64)     | 447 (60)   |
| Female | 199 (44)  | 105 (36)     | 304 (40)   |

**Table 2**
Age distribution across CHD and non-CHD groups.

| Age group   | Age range (Years) | CHD, N (%) | Non-CHD, N (%) | Total, N (%) |
|-------------|-------------------|------------|----------------|--------------|
| Infant      | 1 month to 2      | 125 (27)   | 137 (46)       | 262 (35)     |
| Child       | 2 to 12           | 306 (67)   | 152 (52)       | 458 (61)     |
| Adolescence | 12 to 16          | 22 (5)     | 6 (2)          | 28 (4)       |
| Adult       | > 16              | 3 (1)      | 0 (0)          | 3 (1)        |
| Total       |                   | **456 (61)** | **295 (39)** | **751 (100)** |

**Table 3**
Comparison of weight, height, and BMI between CHD and Non-CHD groups.

|      | Weight (kg) |          | Height (cm) |          | BMI   |          |
|------|-------------|----------|-------------|----------|-------|----------|
|      | CHD         | Non-CHD  | CHD         | Non-CHD  | CHD   | Non-CHD  |
| Max  | 62.5        | 80       | 242         | 177      | 27.34 | 29.6     |
| Min  | 3.4         | 2.3      | 58          | 54       | 4.54  | 4.7      |
| Mean | 15.98       | 15.18    | 103.93      | 98.4     | 14.02 | 14.61    |
| STD  | 8.61        | 10.21    | 24.59       | 26.95    | 2.38  | 3.42     |

*3.2.2. Normalization*

To scale data to a consistent range, we applied z-score normalization to each audio file to ensure consistency and facilitate further analysis by adjusting their mean to zero and their standard deviation to one.

*3.3. Feature extraction*

Various features are extracted from the audio files as described below and illustrated in Fig. 3.

*3.3.1. Handcrafted feature extraction*

To capture both temporal and spectral characteristics of heart sounds, various handcrafted features were initially extracted. Features demonstrating statistical significance based on the Mann–Whitney U test ($p$-value < 0.05) were retained for further analysis. These included heart rate variability metrics, spectral centroid, spectral roll-off, and spectral contrast, which were later combined with deep features for final classification.

- **Heart rate variability (HRV):** These features were calculated using age-based heart rate estimation [40] by monitoring the changes in intervals between each heartbeat, showcasing the dynamic shifts in heart rate over time, which can vary depending on age. Higher HRV metrics, such as the root mean square of successive differences between normal heartbeats (RMSSD) and the standard deviation of NN intervals (SDNN), generally indicate better cardiovascular health and autonomic nervous system function [41].
- **Spectral centroid:** The spectral centroid represents the central point of the power spectrum in an audio signal. It provides information about the "brightness" or "timbre" of the sound, indicating where most of the energy in the frequency spectrum lies [9].
- **Spectral roll-off:** Spectral roll-off refers to the frequency below which a certain percentage of the total spectral energy is concentrated. It characterizes the high-frequency content of the signal and can provide insights into its overall spectral shape [9].
- **Spectral contrast:** Spectral contrast calculates the difference in power levels between the maximum and minimum points in the spectral distribution of an audio signal. It quantifies the perceptual contrast between different frequency bands and can indicate features such as tonality and sharpness in the sound [9].

*3.3.2. MFCC feature extraction*

- **Windowing:** The audio signal is divided into small, 25-millisecond (equivalent to 400 sample points) with 50% overlapping windows. These short segments help manage the non-stationary behavior of the audio signal. Each segment is then multiplied by a window function (the Hamming window), to smooth the edges and minimize spectral leakage, reducing discontinuities and improving Fourier Transform quality.
- **Fourier Transform:** To transform the signal from the time domain to the frequency domain, especially short-time Fourier transform (STFT) is applied to each windowed frame. This step reveals the frequency components within each frame, providing essential information for further processing. For the 25-millisecond window equivalent to 400 sample points, we use 1024 FFT points.

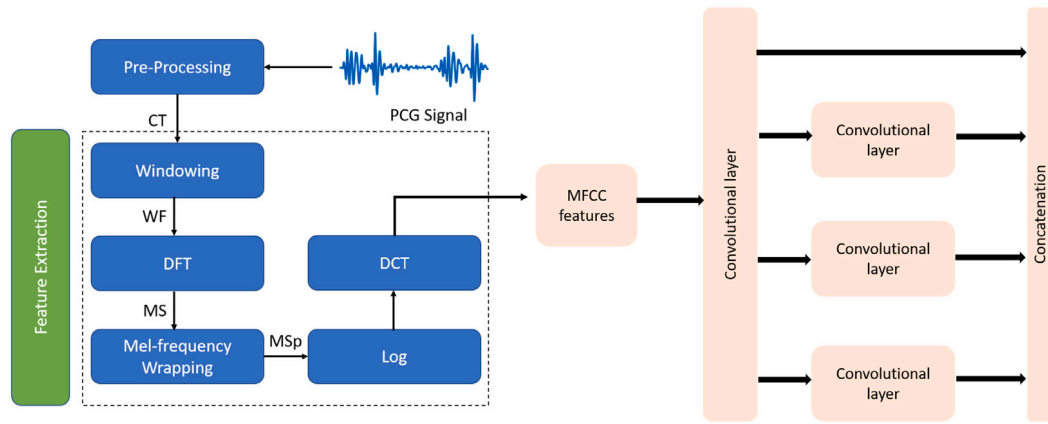

**Fig. 3.** (a) MFCC feature extraction process (b) Architecture of ID CNN (CS: Continuous Signal WF: Windowing Frames MS: Magnitude Spectrum MSp: Mel Spectrum).

- **Mel-frequency warping:** The power spectrum is passed through a mel filter bank. These filters are distributed linearly across lower frequencies and logarithmically across higher frequencies. The mel scale mimics human hearing, which is more sensitive to lower frequencies. This step ensures that the features reflect human auditory perception.
- **Logarithmic scaling:** The energies from each mel filter are converted to a logarithmic scale. It reduces the signal's dynamic range, aligning it more closely with human loudness perception and making the features easier to manage.
- **Discrete Cosine Transform (DCT):** DCT is applied to transform the log filter bank energies to get MFCC features. DCT de-correlates the filter bank energies and emphasizes the most significant coefficients, making the features more suitable for classification tasks. A specified number of coefficients, 13 in this study, are selected from the DCT output. These coefficients capture the most important features of the audio signal's spectral envelope, reducing the data while retaining essential information. The selected MFCC coefficients from each frame are concatenated to form a feature vector.
- **Delta and Delta-Delta Features:** In this stage, the first and second derivatives of MFCCs are computed to capture dynamic changes in the signal, providing further insights into the signal's dynamics.

All calculated features correspond to frequencies within the 25–400 Hz band.

### 3.4. Deep feature extraction with 1D CNN

To enhance the performance of CHD classification, our proposed model leverages a combination of handcrafted and deep features. The overall model architecture is depicted in Fig. 2. We implemented a 1D CNN with four layers to extract deep features. The model takes MFCC as input features. Initially, the MFCC features undergo processing through a convolutional layer. Subsequently, the output from this initial convolutional layer is divided into three parallel convolutional layers as shown in Fig. 3. This branching enables the model to concurrently learn different features, to capture diverse aspects of the sounds. The outputs from these parallel layers are then concatenated, combining them with other handcrafted features into a single vector. This concatenated feature vector is then forwarded to fully connected layers for further processing and final classification. This architectural design allows the model to learn a rich set of features from the input MFCCs, exploiting multiple convolutional layers in parallel, thereby improving performance by capturing a wider variety of patterns and relationships within the data.

**Table 4**
Dataset split and performance metrics for training, validation, and testing sets.

| Dataset | Patients | Samples | Accuracy, % | Sensitivity, % | Specificity, % |
|---|---|---|---|---|---|
| Training | 526 | 2104 | 93.0 | 92.5 | 93.5 |
| Validation | 150 | 600 | 92.0 | 91.5 | 91.0 |
| Testing | 75 | 300 | 92.0 | 91.0 | 92.0 |

### 3.5. Classification

After combining all the features into a single vector array—handcrafted and deep features—these are classified as CHD or non-CHD using a fully connected layer, which consists of a dense layer followed by a softmax activation function.

### 3.6. Training and testing

The proposed model was trained and tested on sound files from 751 individuals. We divided our dataset patient-wise into 70% for training, 20% for validation, and 10% for testing. To mitigate bias, the dataset was partitioned based on patients, ensuring that all signals from the same patient were preserved consistently in the same split dataset throughout the experiment. The CHD classification for each patient was based on averaging the probabilistic outputs of the classifications of sounds recorded from different locations (see Table 4).

### 3.7. Loss function

We use the binary cross-entropy (BCE) loss function for the classification task. BCE is suitable for binary classification, where the goal is to minimize the difference between the predicted probabilities and the true binary labels (CHD vs. Non-CHD). It is defined as:

$$L = -\frac{1}{N} \sum_{i=1}^{N} \left[ y_i \log(p_i) + (1 - y_i) \log(1 - p_i) \right] \tag{1}$$

Where:

- $y_i$ is the true label (1 for CHD, 0 for Non-CHD),
- $p_i$ is the predicted probability of the positive class (CHD),
- $N$ is the total number of samples.

### 3.8. Tackling data imbalance

In the dataset, CHD to non-CHD class ratio is 63:37, indicating a data imbalance. To tackle this issue, we incorporate the class weight function into the loss function. This strategy is crucial for ensuring accurate classification in deep learning algorithms. The weights assigned

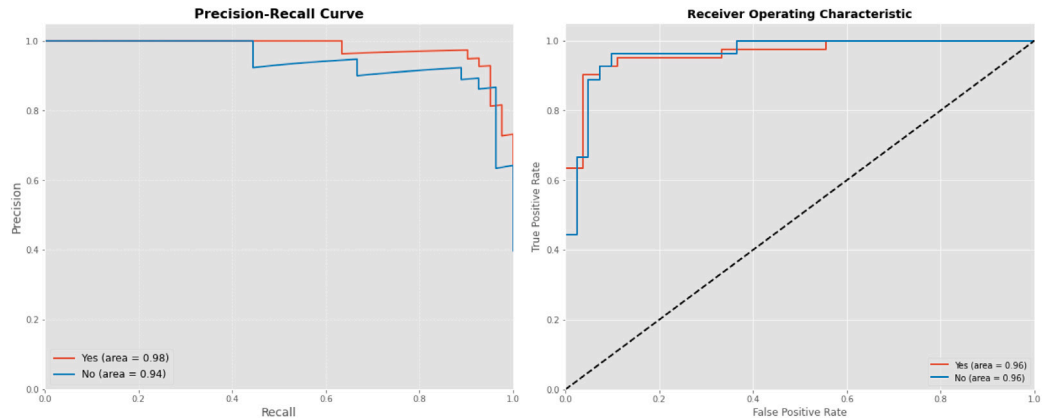

**Fig. 4.** (a) Precision–recall curve (b) ROC curve.

**Table 5**
Feature set selected for the final classification based on the results of the Mann–Whitney U test.

| Feature | p-value |
| --- | --- |
| Mean RR interval (MeanNN) | 0.031 |
| Standard Deviation of NN intervals (SDNN) | 0.012 |
| Root Mean Square of Successive Differences (RMSSD) | 0.008 |
| NN50 (Number of successive RR differences) | 0.045 |
| pNN50 (Proportion of NN50 to total NN intervals) | 0.038 |
| Minimum RR interval | 0.039 |
| Maximum RR interval | 0.024 |
| HRV Triangular Index | 0.016 |
| Spectral Centroid | 0.019 |
| Spectral Roll-off | 0.072 |
| Spectral Contrast | 0.011 |

to the classes were determined based on the equation provided in Eq. (1).

$$W_i = \frac{K}{N} \frac{1}{\sum_{i=1}^{K} a_{ni}} \quad (2)$$

where $W_i$ represents the class weight assigned to each class, $K$ is the total number of samples, and $N$ represents the number of classes. $a_{ni}$ indicates that the $n$th sample belongs to the $i$th class.

*3.9. Performance metrics*

We assessed our model performance using various performance metrics, including accuracy, sensitivity, specificity, AUROC, and Area Under the Precision–Recall Curve (AUPRC).

- **Accuracy**: The proportion of correct predictions made by the model. It is defined as:

$$\text{Accuracy} = \frac{TP + TN}{TP + TN + FP + FN} \quad (3)$$

where $TP$ is the number of true positives, $TN$ is the number of true negatives, $FP$ is the number of false positives, and $FN$ is the number of false negatives.

- **Sensitivity (Recall)**: Measures the model's ability to correctly identify positive cases (CHD). It is defined as:

$$\text{Sensitivity} = \frac{TP}{TP + FN} \quad (4)$$

This metric is crucial for minimizing missed positive cases, especially in medical diagnoses.

- **Specificity**: Measures the model's ability to correctly identify negative cases (non-CHD). It is defined as:

$$\text{Specificity} = \frac{TN}{TN + FP} \quad (5)$$

High specificity ensures that the model correctly classifies healthy individuals as non-CHD.

- **AUROC**: Provides a summary of the model's performance across all classification thresholds. It is the area under the curve plotting the true positive rate versus the false positive rate. A higher AUROC score indicates better classification performance.

These metrics allow for a comprehensive evaluation of the model, balancing the trade-offs between identifying positive cases and minimizing false positives, crucial for detecting CHD in pediatric patients.

*3.10. Feature selection*

To identify features that effectively distinguish between CHD and non-CHD groups, we conducted a statistical evaluation using the Mann–Whitney U test for each candidate feature. Features with a *p*-value < 0.05 were considered statistically significant and included in the model. This approach ensured that only features demonstrating meaningful group differences contributed to model training, improving both performance and interpretability (see Table 5).

**4. Results**

This section presents the performance evaluation of our proposed model for CHD classification based on PCG signals.

*4.1. Overall model performance*

Our primary objective is to evaluate the effectiveness of our model, which relies on multi domain feature fusion for classifying CHD from PCG signals. All results based on the static split are presented in Table 7. The findings indicate that the model exhibits strong performance across all evaluated aspects, including accuracy, sensitivity, specificity, and AUROC. Additionally, Fig. 4 displays the ROC curve and precision–recall curve.

*4.2. Cross-validation-based performance analysis*

To assess the model generalizability and robustness, we performed 5-fold cross-validation using the average probabilistic method. Additionally, we reported the results based on a static data split and compared our model performance with existing state-of-the-art models, as those studies also used a static split for evaluation. The averaged performance metrics from the cross-validation are presented in Table 6 to demonstrate the model consistency across different data partitions.

**Table 6**
Model performance based on 5-fold cross-validation.

| Metric | Accuracy (%) | Sensitivity (%) | Specificity (%) | AUROC (%) |
| --- | --- | --- | --- | --- |
| Mean ± SD | 91.0 ± 2.74 | 90.4 ± 3.34 | 89.5 ± 3.52 | 93.0 ± 2.82 |

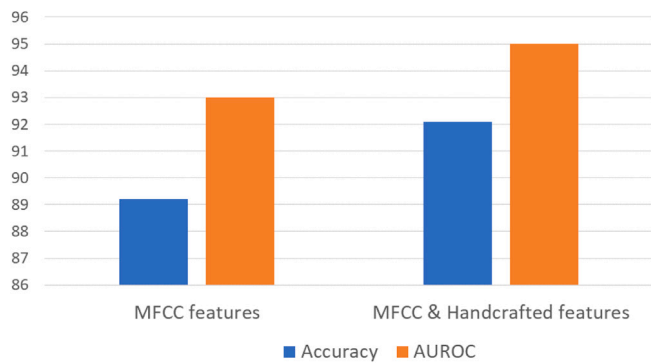

**Fig. 5.** Contribution of each feature to CHD detection.

*4.3. Comparison with existing studies*

As shown in Table 7, our model demonstrates balanced performance across accuracy, sensitivity, specificity, and AUROC, outperforming previous algorithms used for CHD classification in terms of accuracy, specificity, and AUROC. We compared our model's performance with the top two performers of the PhysioNet Challenge 2022 as well as other models developed using the same dataset.

*4.4. Patient-wise decision aggregation methods*

Since we have PCG signals recorded from multiple locations on the heart, we evaluated three different decision aggregation approaches based on multiple PCG recordings from each patient: *at least one positive*, *majority voting*, and *average probabilistic* classification. Table 8 summarizes the results for these methods.

*4.5. Feature contribution analysis*

To assess the contribution of handcrafted and deep features, we measured their impact on CHD detection in terms of accuracy and AUROC. Fig. 5 illustrates these contributions, highlighting the importance of MFCC features.

**5. Discussion**

*5.1. Critical interpretation of results*

The results of our proposed model demonstrate robust and balanced performance, achieving an accuracy of 92%, sensitivity of 91%, specificity of 92%, and AUROC of 96.4%. These results are both statistically significant and clinically meaningful, particularly for pediatric CHD screening. The high specificity of our model is a crucial advancement compared to previous studies, such as those by Xu et al. [33] and Lu et al. [34], whose models yielded lower specificity scores (65.5% and 75.5%, respectively). A lower specificity in screening applications often results in excessive false positives, leading to unnecessary referrals and increased burden on limited clinical resources. In contrast, our model maintains a balance, ensuring both early detection and efficient triage.

It is important to note that for the comparative evaluation, we used the publicly available code implementations provided by the original authors, including those from the PhysioNet 2022 Challenge. This approach ensured consistency and reproducibility of results while minimizing potential discrepancies that might arise from re-implementing the methods. Furthermore, the main focus of the PhysioNet Challenge 2022 is murmur detection, whereas the primary focus of our paper is CHD detection, as murmurs do not always indicate CHD, and some murmurs are clinically innocent. A significant factor contributing to this performance is the integration of deep and handcrafted features, particularly the strong contribution of MFCC features, which aligns with existing literature [42,43] regarding their utility in audio-based medical diagnostics.

Compared to the model proposed by Hassanuzzaman et al. for CHD screening, our approach achieved similar accuracy but offered superior specificity, while sensitivity was slightly lower by 6%. An important consideration is that their model, which relied on a transformer-based architecture trained on raw PCG signals, was developed using a smaller dataset of 440 patients. Processing raw signals directly may increase computational demands, making such models less feasible for deployment in resource-constrained settings. In contrast, our method processed a larger dataset of 751 patients and achieved an average prediction time of 440 ms per 15-second sample on a standard Intel Core i7 CPU, underscoring its practical computational efficiency. Beyond these quantitative improvements, our model introduces a hybrid strategy by integrating both deep features and handcrafted features to enhance the feature representation for better classification. This fusion approach leverages the complementary strengths of data-driven and domain-specific features, enabling our model to deliver competitive or superior AUROC and specificity while remaining computationally lightweight.

As there are four or more sounds from each patient from different locations, we utilized all sounds for CHD disease classification. We conducted experiments with three different approaches: majority voting, where the most common output is the final classification; average probability, where we average all the classification probabilities from the recordings and then classify based on this average probability; and at least one positive, where if at least one recording is classified as positive, the patient is classified as CHD-positive. The results demonstrated that using the greedy approach leads to low specificity, causing the model to predict more non-CHD cases as CHD, thereby increasing the burden on the healthcare system. The majority voting approach resulted in low sensitivity, resulting in more false negatives and may miss identifying CHDs. The comparison in terms of accuracy, sensitivity, specificity and AUROC is provided in Table 8. This table shows that the average probabilistic method demonstrated the best trade-off between sensitivity and specificity. This method avoids the drawbacks of overly aggressive strategies, such as "at least one positive", which can reduce specificity, or majority voting, which can increase false negatives. Therefore, the probabilistic averaging strategy is best suited for clinical applications where both metrics are critical.

We also conducted experiments to assess the contribution of each feature to CHD detection in terms of accuracy and AUROC value. The results are presented in the bar chart in Fig. 5. As depicted in the bar chart, MFCC features made the most significant contribution to CHD detection, whereas handcrafted features accounted for approximately 3%. This is consistent with previous studies that demonstrated their efficacy in biomedical acoustic analysis [42,43]. Additionally, the 1D CNN architecture's multi-path design enables the model to learn diverse features in parallel, improving generalization without the computational overhead of more complex models like transformers. The decision to fuse all variables in our model is based on the understanding that CHD is a complex condition influenced by multiple factors. By integrating diverse types of data, we aim to create a more holistic representation of each case, allowing the model to identify patterns that may not be apparent when considering features individually.

**Table 7**
Performance comparison with existing studies.

| Author (Year) | Dataset | Accuracy, % | Sensitivity, % | Specificity, % | AUROC, % |
|---|---|---|---|---|---|
| Yujia Xu *et al.* (2022) [33] | PhysioNet2022 | 83.8 | 94.4 | 65.5 | 92.2 |
| Hui Lu *et al.* (2022) [34] | PhysioNet2022 | 88.2 | 95.5 | 75.5 | 86.6 |
| Hassanuzzaman *et al.* (2023) [37] | Private | 92 | **97** | 83 | 96 |
| Our Work | Private | **92** | 91 | **92** | **96.4** |

**Table 8**
Performance comparison based on testing method.

| Testing Method | Accuracy, % | Sensitivity, % | Specificity, % | AUROC, % |
|---|---|---|---|---|
| At Least One Positive | 83 | 95 | 71 | 93.3 |
| Majority Voting | 84 | 86 | 81 | 87.4 |
| Average Probabilistic | **92** | 91 | **92** | **96.4** |

*5.2. Clinical relevance*

CHD is a leading cause of infant mortality globally, with higher prevalence in LMICs where access to echocardiography and cardiologists is limited [44,45]. Our model directly addresses this gap by enabling scalable, cost-effective screening using digital stethoscopes. Clinically, our solution enables early identification of CHD in pediatric patients, facilitating timely referral and intervention. The balanced performance of the model ensures that most true cases of CHD are detected (high sensitivity), while unnecessary follow-ups are minimized (high specificity). This is crucial in LMICs, where healthcare resources are scarce and misdiagnoses can have severe consequences. Moreover, the computational efficiency of our approach makes it ideal for deployment in portable devices or mobile applications. This supports its integration into telemedicine platforms, allowing experts to evaluate remotely and extending specialist care to underserved populations. Such capabilities align with global health priorities to expand equitable access to pediatric cardiac screening.

**Limitations leading to future work**

For this study, ECG has also been collected by the Eko device, which could be considered in future research. Moreover, we acknowledge that Our model was trained on a dataset from a single country using one recording device, which may limit its generalizability. Future efforts should focus on incorporating datasets from multiple geographic and device sources.

The inclusion of clinical features and multi-modal approaches in future work may potentially lead to performance improvements. Furthermore, implementing a signal quality assessment module during pre-processing can further improve robustness and reduce the risk of analysis errors due to noisy recordings.

Although this study employed a binary classification setup (CHD vs. non-CHD), expanding to multi-class classification could provide subtype-specific screening in future work. Furthermore, stratified classification based on specific CHD subtypes and patient age could enhance clinical utility. Lastly, additional interpretability techniques could be applied to improve clinician trust in model outputs, aiding in real-world adoption.

**6. Conclusion**

This study introduces a novel approach that utilizes both handcrafted and deep features extracted from PCG sounds, fusing them to enhance classification performance. The deep feature fusion method was chosen for its superior performance in classification compared to traditional deep learning models. This study examines the available records of each patient in the dataset. The deep features, extracted by a 1D CNN, combined with handcrafted, provide a comprehensive representation of heart conditions. The resulting features are then used for classification with a fully connected layer and softmax function. Impressively, the proposed approach achieved an accuracy of 92% for CHD classification, underscoring the robustness of the method as reported in the literature.

**CRediT authorship contribution statement**

**Abdul Jabbar:** Writing – review & editing, Writing – original draft, Visualization, Validation, Software, Resources, Methodology, Investigation, Formal analysis, Conceptualization. **Ethan Grooby:** Writing – review & editing. **Yang Yi Poh:** Writing – review & editing. **Khawza I. Ahmad:** Resources, Formal analysis. **Md Hassanuzzaman:** Resources. **Raqibul Mostafa:** Resources. **Ahsan H. Khandoker:** Writing – review & editing, Resources, Investigation. **Faezeh Marzbanrad:** Writing – review & editing, Writing – original draft, Visualization, Validation, Supervision, Resources, Investigation, Funding acquisition, Formal analysis, Conceptualization.

**Declaration of competing interest**

The authors declare the following financial interests/personal relationships which may be considered as potential competing interests: Abdul Jabbar reports a relationship with Monash University that includes: employment. No, If there are other authors, they declare that they have no known competing financial interests or personal relationships that could have appeared to influence the work reported in this paper.

**Acknowledgments**

This work was supported by a Monash Graduate Scholarship (MGS) from Monash University. This work was partially supported by the Institute for Advanced Research (IAR) of United International University, Bangladesh, under Award UIU/IAR/02/2019-20/SE/10; and in part by the Healthcare Engineering Innovation Group (HEIG), Khalifa University, United Arab Emirates under Award 8474000132.

**References**

[1] M. Hassanuzzaman, N.A. Hasan, M.A.A. Mamun, K.I. Ahmed, A.H. Khandoker, R. Mostafa, Classification of short segment pediatric heart sounds based on a transformer-based convolutional neural network, 2024, arXiv:2404.00470.

[2] Z. Imran, E. Grooby, V.V. Malgi, C. Sitaula, S. Aryal, F. Marzbanrad, A fusion of handcrafted feature-based and deep learning classifiers for heart murmur detection, in: 2022 Computing in Cardiology, CinC, Tampere, Finland, 2022, pp. 1–4, http://dx.doi.org/10.22489/CinC.2022.310.

[3] N.K. Bhat, M. Dhar, R. Kumar, A. Patel, A. Rawat, B.P. Kalra, Prevalence and pattern of congenital heart disease in Uttarakhand, India, Indian J. Pediatr. 80 (2013) 281–285.

[4] S.M. Gilboa, O.J. Devine, J.E. Kucik, M.E. Oster, T. Riehle-Colarusso, W.N. Nembhard, A.J. Marelli, Congenital heart defects in the United States: estimating the magnitude of the affected population in 2010, Circulation 134 (2) (2016) 101–109.

[5] R. Vassar, S. Peyvandi, D. Gano, S. Cox, Y. Zetino, S. Miller, P. McQuillen, Critical congenital heart disease beyond HLHS and TGA: neonatal brain injury and early neurodevelopment, Pediatr. Res. 94 (2) (2023) 691–698, http://dx.doi.org/10.1038/s41390-023-02490-9, Epub 2023 Feb 13.

[6] S. Lopes, I. Guimarães, S. Costa, A. Acosta, K. Sandes, C. Mendes, Mortality for critical congenital heart diseases and associated risk factors in newborns. A cohort study, Arq. Bras. Cardiol. 111 (5) (2018) 666–673, http://dx.doi.org/10.5935/abc.20180175, Epub 2018 Sep 21.

[7] C.A. Warnes, R. Liberthson, G.K. Danielson, A. Dore, L. Harris, J.I. Hoffman, G.D. Webb, Task force 1: the changing profile of congenital heart disease in adult life, J. Am. Coll. Cardiol. 37 (5) (2001) 1170–1175.

[8] W.T. Mahle, J.W. Newburger, G.P. Matherne, F.C. Smith, T.R. Hoke, R. Koppel, S.D. Grosse, Role of pulse oximetry in examining newborns for congenital heart disease: a scientific statement from the American Heart Association and American Academy of Pediatrics, Circulation 120 (5) (2009) 447–458.

[9] S. Abbas, S. Ojo, A. Al Hejaili, et al., Artificial intelligence framework for heart disease classification from audio signals, Sci. Rep. 14 (2024) 3123, http://dx.doi.org/10.1038/s41598-024-53778-7.

[10] Y. Zeinali, S.T.A. Niaki, Heart sound classification using signal processing and machine learning algorithms, Mach. Learn. Appl. 7 (2022) 100206, http://dx.doi.org/10.1016/j.mlwa.2021.100206, URL https://www.sciencedirect.com/science/article/pii/S2666827021001031.

[11] J.-D. Huang, et al., Applying artificial intelligence to wearable sensor data to diagnose and predict cardiovascular disease: A review, Sensors 22 (2022) 8002.

[12] A. Pandey, D. Adedinsewo, The future of AI-enhanced ECG interpretation for valvular heart disease screening, 2022.

[13] H. Kui, J. Pan, R. Zong, H. Yang, W. Wang, Heart sound classification based on log Mel-frequency spectral coefficients features and convolutional neural networks, Biomed. Signal Process. Control. 69 (2021) 102893.

[14] A. Raza, et al., Heartbeat sound signal classification using deep learning, Sensors 19 (2019) 4819.

[15] O. El Badlaoui, A. Benba, A. Hammouch, Novel PCG analysis method for discriminating between abnormal and normal heart sounds, IRBM 41 (4) (2020) 223–228.

[16] H. Wu, S. Kim, K. Bae, Hidden Markov model with heart sound signals for identification of heart diseases, in: Proceedings of 20th International Congress on Acoustics, ICA, Sydney, Australia, 2010, pp. 23–27.

[17] I. Maglogiannis, E. Loukis, E. Zafiropoulos, A. Stasis, Support vectors machine-based identification of heart valve diseases using heart sounds, Comput. Biol. Med. 95 (1) (2009) 47–61.

[18] J. Burns, M. Ganigara, A. Dhar, Application of intelligent phonocardiography in the detection of congenital heart disease in pediatric patients: a narrative review, Prog. Pediatr. Cardiol. 64 (2022) 101455.

[19] P. Chen, Q. Zhang, Classification of heart sounds using discrete time-frequency energy feature based on s transform and the wavelet threshold denoising, Biomed. Signal Process. Control. 57 (2020) 101684.

[20] J. Karhade, S. Dash, S.K. Ghosh, D.K. Dash, R.K. Tripathy, Time–frequency-domain deep learning framework for the automated detection of heart valve disorders using PCG signals, IEEE Trans. Instrum. Meas. 71 (2022) 1–1.

[21] S.K. Ghosh, R. Ponnalagu, R. Tripathy, U.R. Acharya, Automated detection of heart valve diseases using chirplet transform and multiclass composite classifier with PCG signals, Comput. Biol. Med. 118 (2020) 103632.

[22] U. Desai, et al., Decision support system for arrhythmia beats using ECG signals with DCT, DWT and EMD methods: a comparative study, J. Mech. Med. Biology 16 (1) (2016) 1640012.

[23] C. Liu, D. Springer, Q. Li, B. Moody, R.A. Juan, F.J. Chorro, et al., An open access database for the evaluation of heart sound algorithms, Physiol. Meas. 37 (12) (2016) 2181.

[24] M.N. Homsi, F. Plesinger, P. Jurak, L. Viscor, P. Leinveber, I. Halamek, J. Meste, R. Smisek, J.P. Martinek, M. Vondra, Automatic heart sound recording classification using a nested set of ensemble algorithms, in: 2016 Computing in Cardiology Conference, CinC, Vancouver, BC, Canada, 2016, pp. 817–820.

[25] M. Zabihi, A.B. Rad, S. Kiranyaz, M. Gabbouj, A.K. Katsaggelos, Heart sound anomaly and quality detection using ensemble of neural networks without segmentation, in: 2016 Computing in Cardiology Conference, CinC, Vancouver, BC, Canada, 2016, pp. 613–616.

[26] C. Potes, S. Parvaneh, A. Rahman, B. Conroy, Ensemble of feature-based and deep learning-based classifiers for detection of abnormal heart sounds, in: 2016 Computing in Cardiology Conference, CinC, Vancouver, BC, Canada, 2016, pp. 621–624.

[27] H. Li, et al., A fusion framework based on multi-domain features and deep learning features of phonocardiogram for coronary artery disease detection, Comput. Biol. Med. 120 (2020) 103733.

[28] S.A. Singh, T.G. Meitei, S. Majumder, Short PCG Classification Based on Deep Learning, Elsevier, 2020, pp. 141–164.

[29] M. Alkhodari, L. Fraiwan, Convolutional and recurrent neural networks for the detection of valvular heart diseases in phonocardiogram recordings, Comput. Biol. Med. 200 (2021) 105940.

[30] B. Bozkurt, I. Germanakis, Y. Stylianou, A study of time-frequency features for CNN-based automatic heart sound classification for pathology detection, Comput. Biol. Med. 100 (2018) 132–143.

[31] M. Deng, T. Meng, J. Cao, S. Wang, J. Zhang, H. Fan, Heart sound classification based on improved MFCC features and convolutional recurrent neural networks, Neural Netw. 130 (2020) 22–32, http://dx.doi.org/10.1016/j.neunet.2020.06.015, Epub 2020 Jun 23. PMID: 32589588.

[32] M. Reyna, Y. Kiarashi, A. Elola, J. Oliveira, F. Renna, A. Gu, E.A. Perez Alday, N. Sadr, S. Mattos, M. Coimbra, R. Sameni, A. Bahrami Rad, Z. Koscova, G. Clifford, Heart murmur detection from phonocardiogram recordings: The george B. moody PhysioNet challenge 2022 (version 1.0.0), PhysioNet (2023) http://dx.doi.org/10.13026/t49p-5v35.

[33] Y. Xu, X. Bao, H.K. Lam, E.N. Kamavuako, Hierarchical multi-scale convolutional network for murmurs detection on PCG signals, in: 2022 Computing in Cardiology, CinC, Tampere, Finland, 2022, pp. 1–4, http://dx.doi.org/10.22489/CinC.2022.439.

[34] H. Lu, Y. Zhang, Q. Liu, J. Wang, M. Zhou, J. Li, A lightweight robust approach for automatic heart murmurs and clinical outcomes classification from phonocardiogram recordings, in: 2022 Computing in Cardiology, CinC, Tampere, Finland, 2022, pp. 1–4, http://dx.doi.org/10.22489/CinC.2022.165.

[35] A. McDonald, M.J. Gales, A. Agarwal, Detection of heart murmurs in phonocardiograms with parallel hidden semi-Markov models, in: 2022 Computing in Cardiology, CinC, Tampere, Finland, 2022, pp. 1–4, http://dx.doi.org/10.22489/CinC.2022.020.

[36] S. Aziz, M.U. Khan, M. Alhaisoni, T. Akram, M. Altaf, Phonocardiogram signal processing for automatic diagnosis of congenital heart disorders through fusion of temporal and cepstral features, Sensors 20 (13) (2020) http://dx.doi.org/10.3390/s20133790, URL https://www.mdpi.com/1424-8220/20/13/3790.

[37] M. Hassanuzzaman, et al., Recognition of pediatric congenital heart diseases by using phonocardiogram signals and transformer-based neural networks, in: 2023 45th Annual International Conference of the IEEE Engineering in Medicine & Biology Society, EMBC, Sydney, Australia, 2023, pp. 1–4, http://dx.doi.org/10.1109/EMBC40787.2023.10340370.

[38] H.K. Alkhathani, I.U. Haq, Y.Y. Ghadi, N. Innab, M. Alajmi, M. Nurbapa, Precision diagnosis: An automated method for detecting congenital heart diseases in children from phonocardiogram signals employing deep neural network, IEEE Access 12 (2024) 76053–76064, http://dx.doi.org/10.1109/ACCESS.2024.3395389.

[39] O. Tutsoy, M.Y. Tanrikulu, Priority and age specific vaccination algorithm for the pandemic diseases: A comprehensive parametric prediction model, BMC Med. Inform. Decis. Mak. 22 (1) (2022) 4.

[40] E. Grooby, et al., Neonatal heart and lung sound quality assessment for robust heart and breathing rate estimation for telehealth applications, IEEE J. Biomed. Heal. Inform. 25 (12) (2021) 4255–4266, http://dx.doi.org/10.1109/JBHI.2020.3047602.

[41] L. Yugar, J. Yugar-Toledo, N. Dinamarco, L. Sedenho-Prado, B. Moreno, T. Rubio, A. Fattori, B. Rodrigues, J. Vilela-Martin, H. Moreno, The role of heart rate variability (HRV) in different hypertensive syndromes, Diagn. (Basel) 13 (4) (2023) 785, http://dx.doi.org/10.3390/diagnostics13040785.

[42] A.M. Rahmani, A. Haider, M. Adeli, O. Mzoughi, E. Gemeay, M. Mohammadi, H. Alinejad-Rokny, P. Khoshvaght, M. Hosseinzadeh, Enhanced heart sound classification using mel frequency cepstral coefficients and comparative analysis of single vs. ensemble classifier strategies, PLoS One 19 (12) (2024) e0316645, http://dx.doi.org/10.1371/journal.pone.0316645.

[43] M. Deng, T. Meng, J. Cao, S. Wang, J. Zhang, H. Fan, Heart sound classification based on improved MFCC features and convolutional recurrent neural networks, Neural Netw. 130 (2020) 22–32, http://dx.doi.org/10.1016/j.neunet.2020.06.015.

[44] N.K. Bhat, M. Dhar, R. Kumar, A. Patel, A. Rawat, B.P. Kalra, Prevalence and pattern of congenital heart disease in Uttarakhand, India, Indian J. Pediatr. 80 (4) (2013) 281–285, http://dx.doi.org/10.1007/s12098-012-0738-4.

[45] S.M. Gilboa, O.J. Devine, J.E. Kucik, M.E. Oster, T. Riehle-Colarusso, W.N. Nembhard, P. Xu, A. Correa, K. Jenkins, A.J. Marelli, Congenital heart defects in the United States: Estimating the magnitude of the affected population in 2010, Circulation 134 (2) (2016) 101–109, http://dx.doi.org/10.1161/CIRCULATIONAHA.115.019307.